\documentclass[letter,journal]{IEEEtran}
\usepackage{amsmath,amsfonts}
\usepackage{algorithmic}
\usepackage{algorithm}
\usepackage{array}
\usepackage{adjustbox}
\usepackage[normalem]{ulem}
\usepackage[font=footnotesize]{subfig} 
\usepackage[labelsep=period,font=small]{caption}
\usepackage{textcomp}
\usepackage{stfloats}
\usepackage{url}
\usepackage{verbatim}
\usepackage{graphicx}
\usepackage{cite}
\usepackage[table]{xcolor}
\usepackage{tabularx}
\usepackage{soul}
\usepackage{wrapfig}
\usepackage{graphicx}
\newcommand{\fref}[1]{Fig.~\ref{#1}}
\newcommand{\tref}[1]{Table~\ref{#1}}
\newcommand{\sref}[1]{Section~\ref{#1}}

\usepackage[misc]{ifsym}

\begin{document}

\title{Catching Elusive Depression via Facial Micro-Expression Recognition
\thanks{To appear in {\em IEEE Communications Magazine} 2023, under the Feature Topic ``Affective Computing and Communications''.}}

\author{Xiaohui Chen, Tie Luo \textsuperscript{\Letter}, {\it Senior Member, IEEE}\\
Department of Computer Science, Missouri University of Science and Technology, Rolla, MO 65401\\
Email: \{xcqmk, tluo\}@mst.edu}

\maketitle

\begin{abstract}
Depression is a common mental health disorder that can cause consequential symptoms with continuously depressed mood that leads to emotional distress. One category of depression is {\em Concealed Depression}, where patients intentionally or unintentionally hide their genuine emotions through exterior optimism,
thereby complicating and delaying diagnosis and treatment and leading to unexpected suicides.
In this paper, we propose to diagnose concealed depression by using {\em facial micro-expressions} (FMEs) to detect and recognize 
underlying true emotions. However, the extremely low intensity and subtle nature of FMEs make their recognition a tough task. We propose a facial landmark-based Region-of-Interest (ROI) approach to address the challenge, and describe a low-cost and privacy-preserving solution that enables {\em self-diagnosis} using portable mobile devices in a personal setting (e.g., at home). We present results and findings that validate our method, and discuss other technical challenges and future directions in applying such techniques to real clinical settings.
\end{abstract}


\section{Introduction} \label{sec:intro}

\IEEEPARstart{D}{epression}, also known as major depressive disorder or major depression, has been one of the most common mental health disorders both in the United States and globally. It is characterized by persistent sadness and a lack of interest or pleasure in previously rewarding or enjoyable activities, and is a leading cause of disability around the world and contributes greatly to the global burden of disease \cite{who}.
According to the statistics from National Institute of Mental Health (NIMH), in 2020, approximately 21 million or 8\% of all American adults (18+) were diagnosed with at least one major depressive episode and 14.8 million (6\%) had severe impairments~\cite{nimh}. What is equally alarming is that only 65\% of these adults and less than 42\% of adolescent patients received treatment~\cite{nimh}. 


Concealed Depression is one of the major and most frequently occurring depressive disorders nowadays. It is an atypical form of depression where patients intentionally or unintentionally hide their symptoms under exterior optimistic and cheerful behaviors and emotions. Although it can significantly impact patients' quality of living, few patients seek clinical help and, in fact, there is no effective clinical solution to date to diagnose this problem. As a result, this often leads to more cases of delayed or no treatment as compared to major depression, and even suicides eventually. General screening of depression based on patient-administered scoring reports (e.g., CES-D, Zung SDS) is not recommended for asymptomatic cases such as concealed depression due to its subjectivity and lack of scientific proof.

The major challenge in diagnosing concealed depression is that it is extremely subtle to detect and recognize while at the same time patients are not willing to reveal or not even aware of it. In this paper, we propose to diagnose this mental disorder by recognizing {\em facial micro-expressions} (FMEs). An FME is a very subtle and low-intensity change in facial expression, and typically lasts only 1/25 to 1/2 seconds \cite{afma}. It has never been explored for diagnosing concealed depression, where special challenges exist since the exterior and underlying true emotions are contradictory. 

Specifically, this paper makes the following contributions:
\begin{itemize}
  \item This work would create awareness among researchers and practitioners of an overlooked mental disorder, Concealed Depression, due to its increasingly worrying penetration and consequences. We also identify the major challenges that hinder the diagnosis and treatment of this disorder.
    \item We make the first attempt to address this problem using facial micro-expression (FME) recognition as a plausible solution. We propose a novel facial-landmark-based Region-of-Interest (ROI) approach to this FME recognition task. Combined with local feature extraction and machine learning, our approach reveals the strength of correlations between various human emotions and different facial regions (individual and combined).
    \item Our study suggests the potential of a privacy-preserving self-diagnosing method to treat concealed depression, as well as a more viable way to create open datasets for FME analysis and research. We also point out current limitations and possible future research directions on this subject.
\end{itemize}

\section{Background}\label{sec:expr}
The universal definition of facial expressions can be categorized into happiness, surprise, contempt, disgust, sadness, anger and fear \cite{ufe}. Facial Action Coding System (FACS) \cite{facs} first introduced by Carl-Herman Hjortsjö, 
is a coding system that quantifies and measures different facial actions based on muscle movements. Researchers utilize this tool for facial expression and emotion annotation as it provides a standardized and objective way of identifying and measuring specific muscle movements. For example, in the process of creating our chosen dataset CASME-II \cite{casme2}, facial actions AU6 (Cheek Raiser) and AU12 (Lip Corner Puller) are treated as signs for happiness. In \cite{fee}, P. Ekman raised a profound question: can we really rely on facial expressions to identify human emotions? In other words, are facial expressions subject to other potential factors that limit their capability in describing human emotions? 
According to Ekman's paper \cite{fee}, facial expressions can be misleading in interpreting emotions. For example, a person could intentionally show a facial expression that is generally interpreted as a particular emotion while they in fact do not have that emotion. Conversely, people who are facially inactive may not wear any facial expression (either intentionally or unintentionally) while they in fact are experiencing an emotion.

Unlike facial expressions which are "macro-expressions", a micro-expression can reveal genuine human emotions as it is a spontaneous expression of feelings even if one tries to hide or repress his/her true emotion. Micro-expressions were first discovered by Haggard and Isaacs in 1966 from scanning motion pictures during psychotherapy hours, and later re-discovered by Ekman and Freisen in 1969 from a filmed interview of a student who was trying to hide her suicidal intention. We contend that FME represents a unique and promising approach to diagnosing concealed depression because FME is a biological spontaneous process that can hardly be faked for the vast majority of people, regardless of their cultural  and linguistic backgrounds. This is another compelling reason that motivated our study.

{\bf Literature Review.}
Different depression detection techniques were proposed such as sentiment analysis, facial expression analysis, and combination of chatbots and emotional AI, as reviewed by \cite{joshi2022materials}. Recent advances in micro-expression analysis have shown effectiveness in various fields including business negotiation, national security and criminal justice \cite{afma}.
Traditional approaches rely on engineered features that describe changes between video frames using (mainly SVM-based) linear or nonlinear models (not both) with preprocessing (e.g., linear interpolation, optical flow) to recognize FMEs, achieving an overall model accuracy between 45.9-70.3\% on CASME-II dataset \cite{verma2022deep}. Recently, deep learning based approaches were exploited \cite{verma2022deep,li2022ieeetransactions}. For example, Lee et al. \cite{lee2022diagnostics} proposed a CNN based recognition system that combines real-time analysis on speech-dependent questionnaire response and facial images to predict depression. Other studies also explored sequential models such as RCNN and CNN-LSTM. However, our work made the first attempt in relating MER with an overlooked mental disorder ``concealed depression'' where depression is much harder to recognize. In addition, unlike existing approaches which primarily focus on image frames of whole faces, our work investigates different facial regions and combination of upper and lower facial ROIs that brings both lower computation cost and privacy protection. Moreover, our study does not just address model performance but also take into account model efficiency and deployment considerations towards more practical adoption (e.g., self-diagnosing apps).

\begin{figure*}[t]
    \centering    \includegraphics[width=0.75\textwidth]{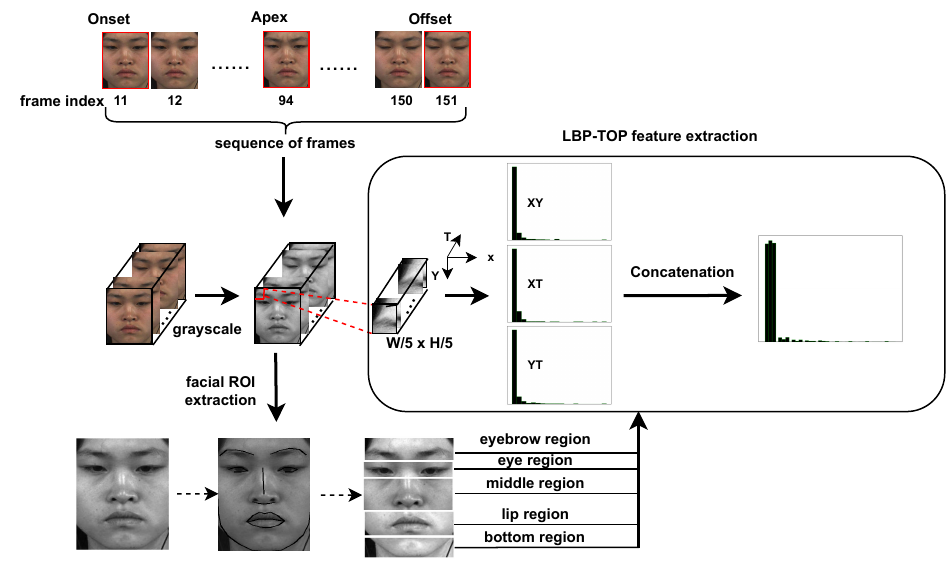}
    \caption{Our preprocessing pipeline for an example video frame sequence. Division factor is 5 for illustration.}
    \label{preprop-diagram}
\end{figure*}

\section{Recognition of Micro-Expressions}
Depression can be seen as an {\em emotional state}. Early psychological research \cite{Abroms1981} discovered characteristics of patients with hidden depression from various case studies that soft signs of mood disorder such as pessimism and low self-esteem contribute to depression recognition, and denying behaviors may also be important indicators to depression with the purpose of hiding internal fear. Majority of the cases show a strong correlation between underlying depression and (usually negative) emotional activities. Interestingly, a later study \cite{Surguladze2010} discovered more sensitive neurobiological response to disgust facial expressions in patients with depression, revealing an overlooked but noteworthy connection between disgust emotion and depression.

In this paper, we propose a method that discovers inference patterns between facial micro-expressions and human emotions for detecting concealed depression from facial image sequences. We aim to achieve two goals. First, to detect true emotions from FMEs as they are spontaneous and involuntary physiological reaction. Second, to determine whether there are depressive signs through recognition of negative emotions. We utilize a pixel-wise feature extraction method in a three-dimensional fashion on both whole face area and facial feature landmark areas, and thereby obtain information on both local texture change as well as temporal pattern in image sequences for different facial areas. In addition, we explore correlation between different facial regions and human emotions, for which we particularly examine the feasibility of recognizing negative emotions through only local facial areas (e.g., eyebrow) rather than whole faces. Note that this has a significant implication on {\em privacy protection}: our study shows that it is possible to diagnose concealed depression without revealing a person's identity.

\subsection{Facial ROI Extraction using Landmarks}

We use image sequences extracted from videos to detect miniature facial movements. In a video clip, each frame sequence has its onset, apex and offset frames annotated as well as its observed facial action units (e.g, upper lid raiser; cf. \sref{sec:expr}). We observe that facial actions can play an important role in emotion recognition. Hence in this study, we divided each facial image into five region-of-interests (ROIs): eyebrow, eye, middle, lip and bottom, where middle ROI consists of the cheek and nose area, and bottom ROI consists of the jaw and chin area. In order to extract these facial ROIs for micro-expression recognition, we employed {\em facial landmark detection} which allows us to locate different facial components. To this end, we first obtain different face landmark locations using Python dlib face landmark detection model based on Histogram of Oriented Gradients (HoGs), and then determine the upper and lower boundaries of each ROI. 

For each frame sequence in a video clip, we want to extract 5 ROIs for each frame. Despite changes between frames, we note that {\em facial landmark locations} remain invariant across frames. As all the frames in a sequence have the same size, we can therefore exploit the facial landmarks from the onset (beginning) frame, which is used as a ``registration'' image, and use those landmarks as reference points to further extract ROIs in subsequent frames up to the offset (stop) frame. Thus, for each image sequence, we construct five sub-sequences using the five extracted facial ROIs. Then, we pass each sub-sequence of images to our feature extraction phase separately, described next.

\subsection{Feature Extraction}

In our pilot study, we used single apex frame from each image sequence to extract useful spatial features. The apex frame, as the name suggests, contains the most information in a facial event or action during a short time duration. We employed basic Local Binary Pattern (LBP) \cite{lbp} which is an efficient texture feature descriptor that has been widely used in tasks such as facial expression recognition. Its computational simplicity and robustness in describing local spatial patterns and grayscale-level changes has made it popular in this field. In LBP, with predefined radius and number of neighboring pixels, we can obtain the binary pattern of each central pixel in an image by thresholding all of its neighboring pixels. Given a radius of 1, we can then obtain a binary pattern through comparison between intensity value of central pixel and each of its 8 neighboring pixels - 1 if the neighboring pixel is greater than the central pixel, otherwise 0. Then, by converting from binary to decimal, we obtain a number to represent the central pixel. Thus, by repeating the above procedure for every pixel in an image, we obtain a histogram of the image by counting the occurrence of each possible decimal value.

However, with LBP we were only able to achieve 43\% accuracy in classifying all the emotions. We identified that the reason was that spatial features itself is only effective for detecting emotions that can be recognized from obvious facial expressions (e.g., eyebrow raise and eye opening is a clear indicator of surprise). In contrast, micro-expressions have much lower intensity and are much more subtle, thereby rendering LBP much less effective. 

Therefore, we take into account an additional dimension that represents temporal features within each sequence of frames. To this end, we use Local Binary Pattern on Three Orthogonal Planes (LBP-TOP) proposed by \cite{lbptop}, to extract spatial-temporal features from image sequences (videos). In LBP-TOP, we need to perform LBP operation on each of the XY, XT and YT planes, where T stands for the time axis, via which we obtain three histogram features and then concatenate them to form the final image feature map. A histogram is a type of feature descriptor representing distribution of all possible pixel values in an image. LBP-TOP brings feature description from two-dimensional to three-dimensional space and enhances image analysis that heavily depend on temporal pattern extraction.

In our study, we partition each image sequence using a 5x5 or 10x10 grid and then extract features using LBP-TOP. A 5x5 partition divides each image
in a sequence into a grid of 5x5 blocks, each of size (height/5, width/5). Then, after extracting LBP-TOP features from each of the 25 blocks, we convert the features into a histogram for each block as follows. Since each pixel (except for those at the boundary of an image) in a block has 256 possible BP patterns ranging from 0 to 255, where 255 corresponds to the binary pattern 11111111 (assuming 8 neighboring pixels), we count the occurrence of each of the 256 values to construct a 256-bin histogram. As these need to be done for each of the XY, XT and YT planes, we end up with 19200 features (flattened) for each sequence (the images sequence along the T dimension was superimposed to construct each histogram, and the 25 histograms are then concatenated).
The above applies similarly to the 10x10 partition, which leads to 76800 features. After that, we perform normalization in view of the large variance in feature values.

\subsection{Algorithms}
Through the feature extraction process, we obtained histogram-like features based on value occurrences. Given the low-level histogram features extracted from each image sequence, we use both linear and nonlinear machine learning algorithms to train our micro-expression recognizer (classifier). For linear models, we use Linear SVM and Logistic Regression classifiers; for non-linear models, we use Random Forest and K-Nearest-Neighbor (KNN) classifiers. We choose them for the following reasons. SVM is well-known for its ability to solve both linear and nonlinear problems by finding an optimal hyperplane to distinguish between different classes with sophisticated data transformation. Linear kernel SVM is especially useful when dealing with a large feature set, which in our case, has 19,200 and 76,800 features. Logistic regression is powerful in analyzing correlation between dependent and independent variables. In our case, we have categorical emotion classes to which a logistic regression classifier can apply; also, its probabilistic nature and fair resistance to overfitting make it a good candidate as well. Random forest is an ensemble learning model that builds upon decision trees and uses different sampling strategies to reach a final decision. It is efficient in producing classification results for large datasets and capable of dealing with imbalanced datasets. KNN as a simple non-parametric learning model uses a distance measure between points in a space to determine similarity. Although it can be computationally costly, the output can sometimes be very accurate and reliable by virtue of its natural, distance-based similarity measures.

\section{Experiments}

\subsection{Dataset}
One of the biggest challenges in facial micro-expression recognition tasks is the limitation of available databases. There are mainly two types of micro-expression databases: posed and spontaneous. For posed micro-expression collection, participants are asked to express various emotions for given tasks. On the other hand, in spontaneous micro-expression collection, participants are asked to hide their feelings and show neutral face, while in the meantime, emotion-inducing movies are played in front of participants and videos are recorded. This method is to ensure a successful induction of spontaneous micro-expressions. Clearly, spontaneous databases mirror the real scenarios more closely and would serve detection and recognition of hidden genuine emotions better. Therefore, in this paper, we use a spontaneous micro-expression database, CASME-II \cite{casme2} (accessing the dataset requires a license agreement which we have signed with the authors). This dataset contains two types of micro-expressions: those when the subjects are not self-aware and those when they intend to suppress their facial movements. There are 255 micro-expression video clips in total of 26 participants, which were preprocessed, cropped, and labeled with seven estimated emotions and their corresponding sample sizes: happy(32), surprise(28), disgust(63), sad(4), fear(2), repression(27), and others(99). The emotions are estimated by using not only participants’ self-report but also facial action units as well as the emotional content of each video that was played. The video recordings were collected using a 200 fps high-speed camera with additional illumination assistance to enhance capturing the facial area of each subject.

\subsection{Preprocessing}

\fref{preprop-diagram} depicts our preprocessing workflow. Image pre-processing is a common practice in computer vision to benefit feature extraction, where typical techniques include denoising, graying, enhancement, registration, normalization, etc. However, in this study, we chose not to apply denoising and enhancement and similar processes as they tend to lose very fine and subtle image details which, however, are essential to FME recognition. On the other hand, in consideration of the relatively small size of the available dataset, we grouped disgust, fear, and sadness into a new category named Negative emotions, and relabeled happiness as Positive emotion; Surprise and Others remain unchanged which are the other two categories. Negative emotions are considered as our target class in the study. For each subject, we extracted every frame sequence that corresponds to an FME event, from onset to apex to offset, after which we converted all RGB images into grayscale.

\subsection{Setup}

In our experiment, we considered the five facial ROIs (eyebrow, eye, middle, lip and bottom) to extract features with two block division factors 5 and 10, which correspond to a flattened feature size of 19200 and 76800, respectively. To explore potentially more effective options, we also combined facial ROIs which eventually leads to nine facial areas: 1) Whole face, 2) Eyebrow ROI, 3) Eye ROI, 4) Middle ROI, consisting of nose and cheek, 5) Lip ROI, 6) Bottom ROI, consisting of jaw and chin, 7) Eyebrow+Eye ROI, 8) Eyebrow+Lip ROI, 9) Eyebrow+Eye+Lip ROI. For each combination of ROIs, we
concatenated their features. After extracting the features, we fed them into each of the four classification models: Linear SVM, Random Forest, KNN and Logistic Regression. We also performed hyperparameter tuning for each model using grid search to optimize their performance, and applied stratified sampling on our dataset due to its imbalanced nature.

We use both accuracy and one-vs-all ROC AUC as our performance metrics, as well as confusion matrix to examine specific class labels. ROC curve allows us to visualize the degree of separability of a model in distinguishing between different classes, where a curve that leans more towards the top left corner indicates better performance with a higher AUC score. 

\begin{table*}[!ht]
\centering
\begin{adjustbox}{width=\textwidth}
\begin{tabular} { 
  | l | c | c | c | c | c | c | c | c | c | c | }
 \hline
  Model & Whole Face & Eyebrow & Eye & Middle & Lip & Bottom & Eyebrow+Eye & Eyebrow+Lip & Eyebrow+Eye+Lip & Mean Accuracy \\ 
 \hline
  \cellcolor{gray!25}avg. CC (5x5) & \cellcolor{gray!25}5.68s & \cellcolor{gray!25}1.1s & \cellcolor{gray!25}0.56s & \cellcolor{gray!25}1.53s & \cellcolor{gray!25}0.96s & \cellcolor{gray!25}0.79s & \cellcolor{gray!25}1.86s & \cellcolor{gray!25}2.35s & \cellcolor{gray!25}3.28s & \\ 
  \hline
  \cellcolor{gray!25}avg. CC (10x10) & \cellcolor{gray!25}5.44s & \cellcolor{gray!25}0.84s & \cellcolor{gray!25}0.31s & \cellcolor{gray!25}1.31s & \cellcolor{gray!25}0.79s & \cellcolor{gray!25}0.56s & \cellcolor{gray!25}1.43s & \cellcolor{gray!25}1.95s & \cellcolor{gray!25}2.57s & \\ 
 \hline
 LSVM5x5 & 62.75\% & 60.78\% & 62.75\% & 56.86\% & 56.86\% & \cellcolor{cyan!30}60.78\% & \cellcolor{cyan!30}66.67\% & 66.67\% & \cellcolor{red!15}70.59\% & 62.75\% \\ 
 \hline
 RF5x5 & 62.75\% & \cellcolor{cyan!30}{\textbf{68.63\%}} & 58.82\% & \cellcolor{cyan!30}60.78\% & \cellcolor{cyan!30}62.75\% & 56.86\% & \cellcolor{cyan!30}66.67\% & \cellcolor{red!15}{\textbf{70.59\%}} & \cellcolor{red!15}{\textbf{70.59\%}} & \color{red}{\textbf{64.27\%}} \\
 \hline
 KNN5x5 & 64.71\% & 60.78\% & \cellcolor{cyan!30}63.07\% & 58.82\% & 60.78\% & 54.90\% & 60.78\% & \cellcolor{yellow!60}66.67\% & 64.71\% & 61.69\% \\
 \hline
 LR5x5 & 60.78\% & 62.75\% & 62.75\% & 54.90\% & 58.82\% & 56.86\% & \cellcolor{yellow!60}64.71\% & 60.78\% & \cellcolor{yellow!60}64.71\% & 60.78\% \\
 \hline
 LSVM10x10 & 64.71\% & 60.78\% & 52.93\% & 58.82\% & 60.78\% & 49.02\% & 62.75\% & \cellcolor{yellow!60}68.63\% & 66.67\% & 60.57\% \\
 \hline
 RF10x10 & \cellcolor{cyan!30}66.67\% & 64.71\% & 56.86\% & 58.82\% & 60.78\% & 58.82\% & 64.71\% & \cellcolor{yellow!60}68.63\% & 60.78\% & 62.31\%\\ 
 \hline
 KNN10x10 & 62.75\% & \cellcolor{yellow!60}64.71\% & 52.94\% & 54.90\% & 58.82\% & 56.86\% & 58.82\% & 56.86\% & 58.82\% & 58.39\%\\
 \hline
 LR10x10 & \cellcolor{cyan!30}66.67\% & 58.82\% & 58.82\% & 50.98\% & 56.86\% & 49.02\% & 62.75\% & \cellcolor{yellow!60}68.63\% & \cellcolor{yellow!60}68.63\% & 60.13\% \\
 \hline
  Mean Accuracy & 63.97\% & 62.75\% & 58.62\% & 56.86\% & 59.56\% & 55.39\% & 63.48\% & \color{red}{\textbf{65.93\%}} & 65.69\% & \\
 \hline
\end{tabular}
\end{adjustbox}
\caption{Comparison of classification accuracy among different models. Yellow cells indicate the 
best-accuracy ROI (horizontally) for each model, blue cells indicate the best-accuracy model (vertically) for each ROI, and pink cells indicate both horizontally and vertically the best performance combination. CC represents computational cost, calculated as the execution time for feature extraction per sample.}\label{demo-table}
\end{table*}

\subsection{Results and Discussion}
Our problem is a multi-class classification problem with a large feature space. We investigate which linear or nonlinear classifiers perform well, and which facial ROIs or combination of ROIs are the most predictive of specific emotion classes.

The results are summarized in \tref{demo-table} where each model has two versions depending on the block division factor (5 or 10). First by looking at the ``Mean accuracy'' for overall performance, we see that the Eyebrow+Lip ROI has the highest mean accuracy (65.93\%) across all models, and Random Forest performs the best across all facial ROIs including whole face, with an accuracy of 64.27\% when the division factor is 5. We attribute this to RF's strong ability of handling nonlinearity which our problem heavily involves. 


Next, by looking at the table vertically, we see that for {\em Whole Face}, the best performers appear at the higher division factor (i.e., 10). RF and LR classifiers both achieve the highest accuracy and, as the division factor increases, the largest performance improvement (from 62.75\% to 66.67\% for RF and from 60.78\% to 66.67\% for LR). The possible reason is due to RF's inherent feature selection ability when the feature dimension is high, and that LR can cope well with overfitting by virtue of regularization (we use Ridge) when the feature space is large. However, for other columns that investigate ROIs instead of whole face, the lower division factor generally yields better performance. This suggests that a moderate resolution of each block is preferable to produce good detection results. Horizontally across the table, it is apparent that combined ROIs perform better than individual ROIs and even Whole Face. In particular, the Eyebrow+Lip ROI is the best performer overall, with the highest accuracy achieved (70.59\%) along with RF5x5.

In summary, combining ROIs from both upper and lower facial parts (e.g., eyebrow and lip) seems to be a good choice and random forest with lower division factor would fare well due to its nonlinearity handling and feature selection capabilities. Our study points out the potential viability of using just the eyebrow+lip ROI in FME recognition, which not only reduces computational cost due to the much fewer input pixels as compared to whole faces, but more importantly is also privacy-preserving. One of the biggest concerns in facial image related applications is privacy, and our ROI-based approach offers a potential solution. Specifically, the ROI extraction procedure can be performed in a private setting (e.g., at home with a mobile app or in a private clinic), and only the ROIs will be sent out for diagnosis or research purposes (e.g., developing a dataset that contains ROI only instead of full facial images, unlike the current practice). In fact, using full facial images is the main reason why the availability of FME datasets has been so limited, and our ROI-based approach could contribute to mitigating this issue.

\begin{figure*}[!ht]
\centering
\subfloat[Eyebrow ROI]{\includegraphics[width=0.25\textwidth]{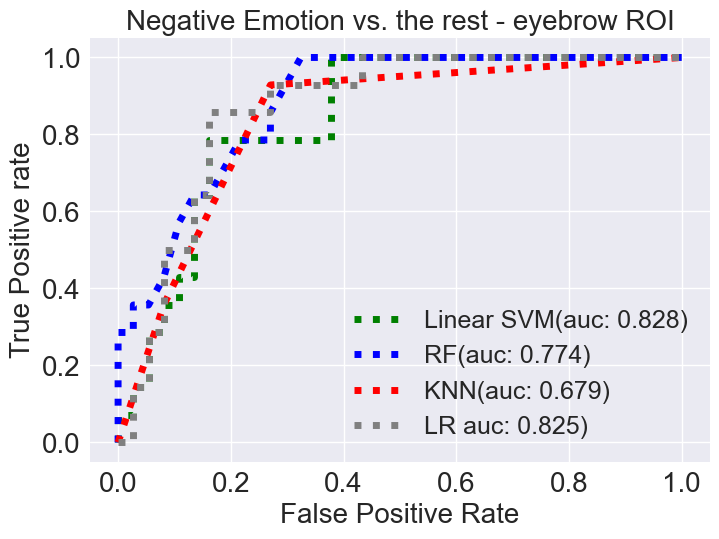}%
\label{fig3.1}}
\hfil
\subfloat[Eye ROI]{\includegraphics[width=0.25\textwidth]{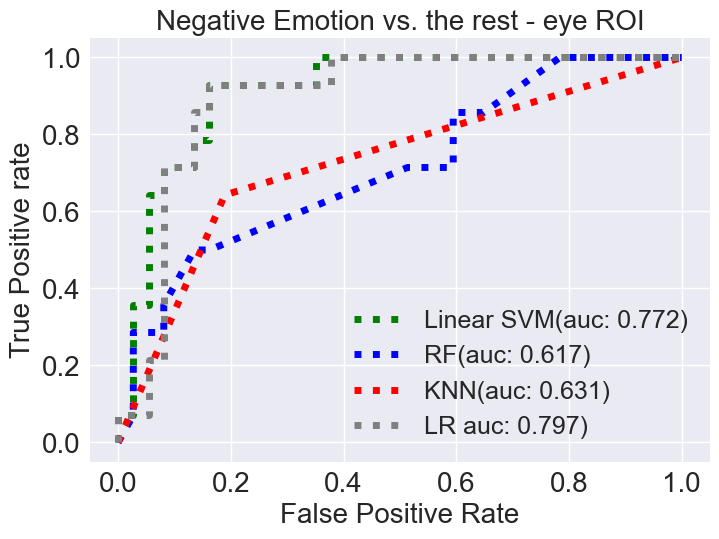}%
\label{fig3.2}}
\hfil
\subfloat[Lip ROI]{\includegraphics[width=0.25\textwidth]{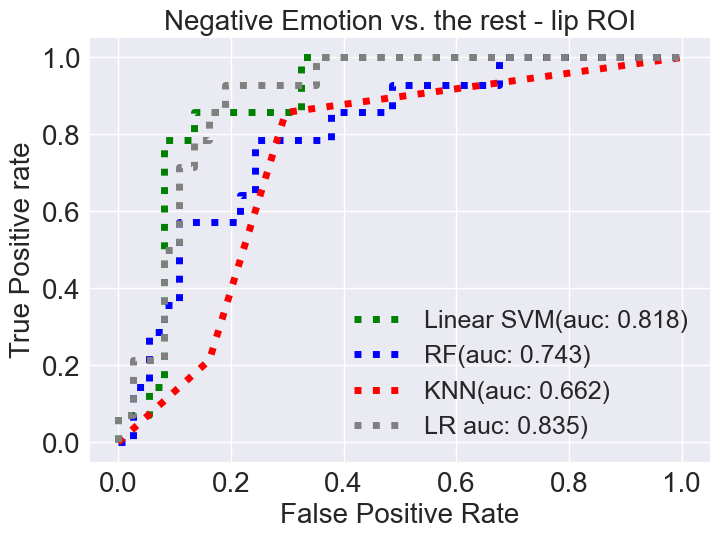}%
\label{fig3.3}}
\hfil
\subfloat[Bottom ROI]{\includegraphics[width=0.25\textwidth]{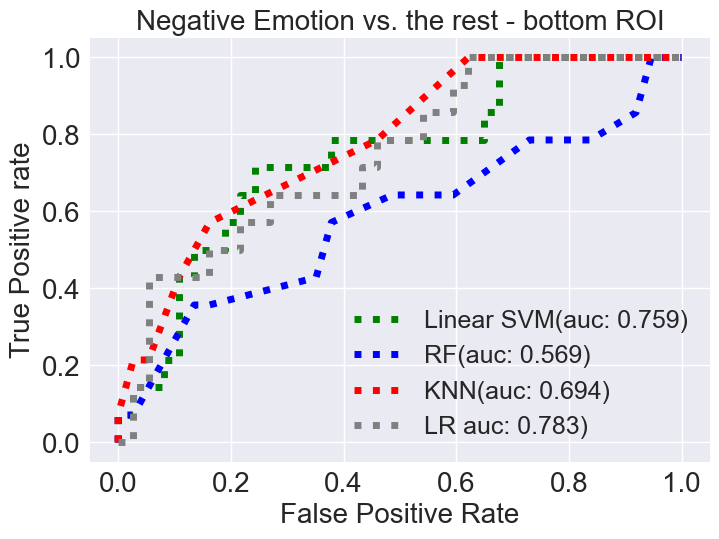}%
\label{fig3.4}}
\caption{ROC Curves for different facial ROIs in recognizing Negative emotions. Each AUC score (in legends) is averaged over all classes. Block division factor is 5.}
\label{fig:ebROC}
\end{figure*}

\begin{figure*}[!ht]
\centering
\subfloat[Eyebrow+Eye+Lip ROI (LSVM)]{\includegraphics[width=0.28\textwidth]{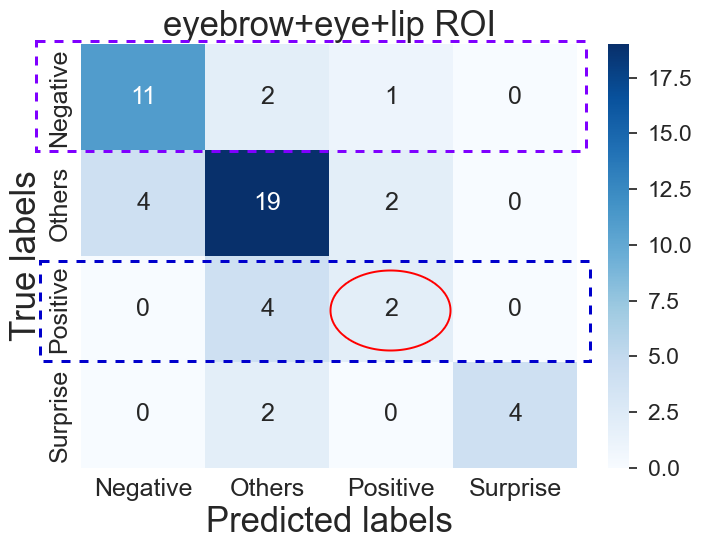}%
\label{fig4.0}}
\hfil
\subfloat[Eyebrow+Lip ROI (LSVM)]{\includegraphics[width=0.28\textwidth]{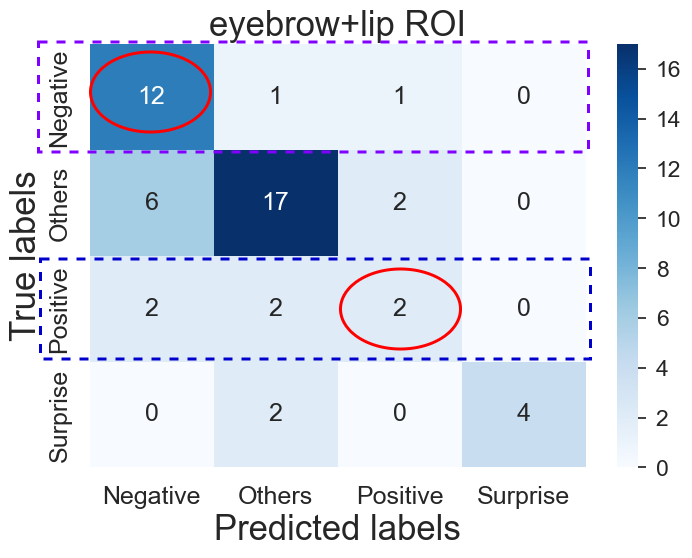}%
\label{fig4.1}}
\hfil
\subfloat[Eyebrow ROI (RF)]{\includegraphics[width=0.28\textwidth]{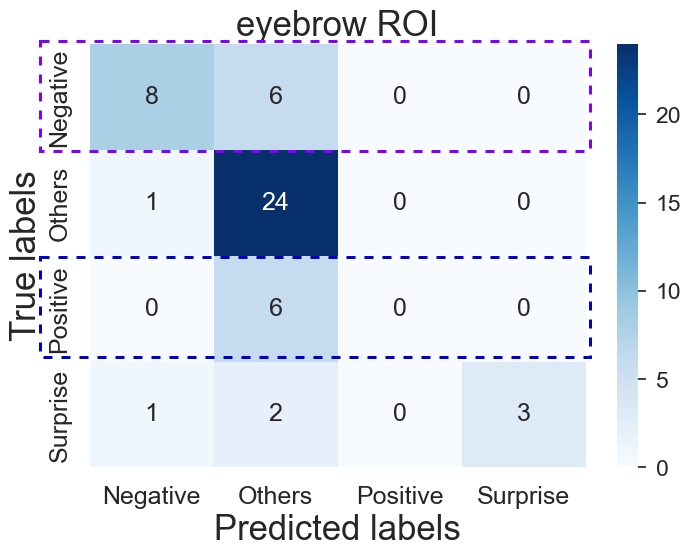}%
\label{fig4.2}}
\hfil
\subfloat[Eye ROI (LR)]{\includegraphics[width=0.28\textwidth]{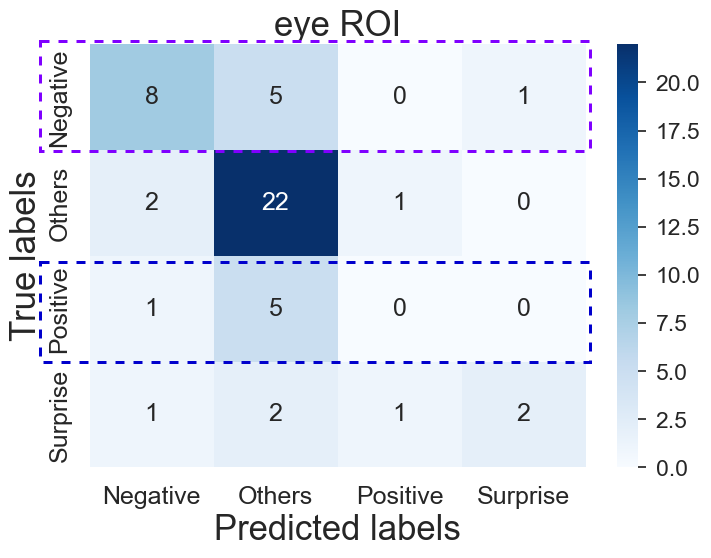}%
\label{fig4.3}}
\hfil
\subfloat[Lip ROI (RF)]{\includegraphics[width=0.28\textwidth]{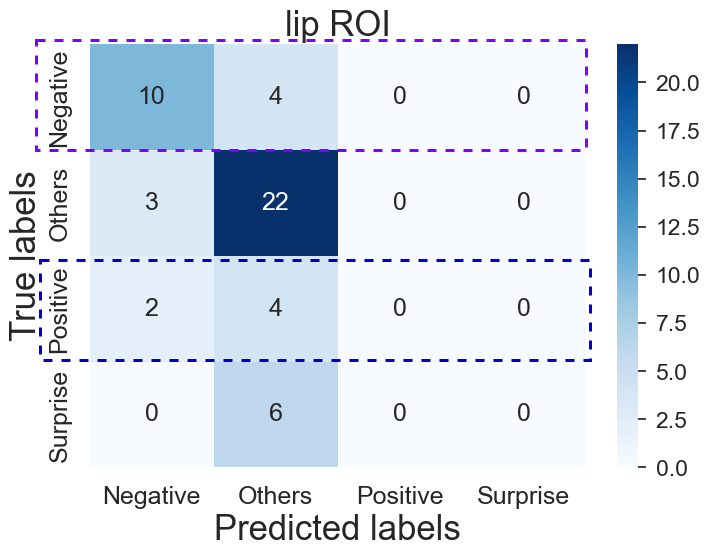}%
\label{fig4.4}}
\hfil
\subfloat[Bottom ROI (LSVM)]{\includegraphics[width=0.28\textwidth]{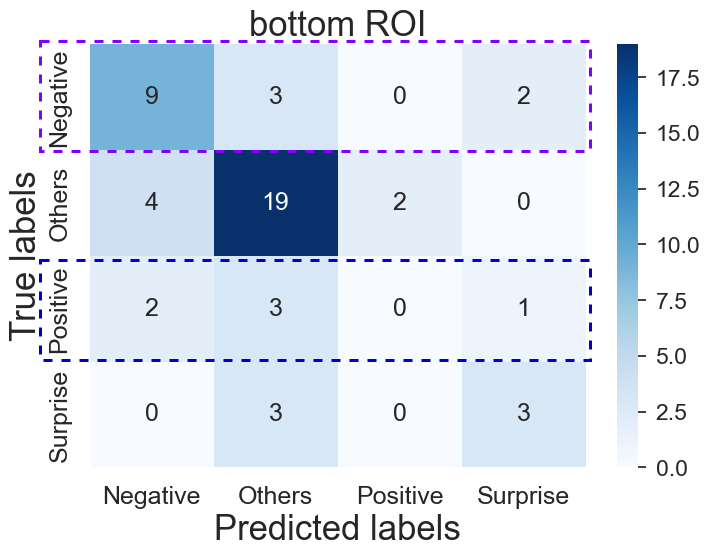}%
\label{fig4.5}}
\caption{Comparison of confusion matrices  with respect to different ROIs. Block division factor is 5.}
\label{fig:faceROI}
\end{figure*}

From the ROC curve plots in \fref{fig:ebROC}, we can see that eyebrow ROI compared to other facial ROIs generates uniformly good results for Negative emotions across all models. We also noticed that none of the facial ROIs can individually perform well in detecting Positive emotion whereas combination of ROIs offers better classification results. This can be seen from our confusion matrix comparison results in \fref{fig:faceROI}.
Our target class for concealed depression detection is Negative emotion, which subsumes sad, disgust and fear. As shown in \fref{fig:ebROC}a, all our models have exhibited effectiveness in distinguishing them from other emotion classes. More specifically, we observed that the eyebrow+lip ROI is effective on this. See \fref{fig:faceROI}b where 12 out 14 Negative emotion cases are correctly predicted. Interestingly, this finding also hints toward the importance of considering both upper and lower facial areas together, as the distinguishable muscle movements from these two areas appear to contribute most to recognizing both Negative and Positive emotions in a collaborative manner.

While FME analysis has yet to be adopted for clinical diagnosis of depression due to technical and ethical concerns, we believe more private or personal applications such as at-home self-diagnosis and monitoring systems can significantly facilitate timely treatment. For example, a smart-home system with a high-resolution camera can capture user’s face at high frame rate and performs FME analysis to detect and recognize user’s implicit emotions locally. 
Since depression generally lasts for a long term (from weeks to years), such a system can continuously monitor user’s emotions to provide reliable feedback on her depression status. Moreover, our experiment results have shown that the combined eyebrow and lip region is representative in FME analysis, and this finding can be exploited in real system implementation to further preserve user privacy.

The property of being lightweight and making near real-time inference makes our model an ideal candidate for deployment on edge devices in self-diagnosis-like applications. Our preprocessing pipeline and training are done offline. After deployment, the costliest step in the workflow is feature extraction. Our experiment was conducted on a 64GB i7-12700H laptop. To extract LBP-TOP features, we utilized numba.jit compilation to execute code in an accelerated CPU mode. To that end, we have empirically shown (cf. \tref{demo-table}) that our proposed facial ROI feature combinations (e.g., eyebrow+lip) instead of whole face are effective in recognizing emotions from FMEs with only slight increase in computational cost (e.g., eyebrow+lip yields cost of 2.35s vs eye with least cost of 0.56s); the corresponding feature extraction as well as other preparatory steps on combined ROIs for each image sequence are computationally inexpensive and take only a few seconds. Moreover, our model size is rather small (the best-performing RF model consists of 25 trees among which even the deepest tree is of depth 14 only), making it feasible for edge deployment.

\section{Conclusion and Future Directions}

Fast-paced life nowadays has brought forth unprecedented social, occupational and emotional pressure and challenges to people. Depression as a result has been constantly on a worrying rise, while unfortunately, many patients are either unaware or not willing to seek professional aid. In this paper, we address an emerging and special type of such mental disorder called concealed depression, which is more challenging than general depression because those patients even try their best to hide their symptoms, letting alone seeing doctors.

We identify facial micro-expressions (FMEs) as a ``decoder'' of humans' hidden emotions, and propose a computer vision based approach to recognize FMEs in order to reveal true feelings or emotions and thereby diagnose concealed depression. Our novel approach consists of facial landmark-based ROI extraction rather than conventional whole face-based emotion detection - our study reveals that eyebrow+lip ROI performs the best in terms of the overall accuracy and has strong capability of detecting Negative emotions, and preserves patients' privacy which has been a long-standing barrier that prevents many people from early diagnosis. Moreover, we employ spatial-temporal feature extraction and lightweight machine learning algorithms; thus, together with our efficient ROI and feature extraction procedures, our proposed approach can be deployed on portable mobile devices and enable self-diagnosis by patients in a personal setting (e.g. at home), which not only maximally protects privacy but also would significant promote the rate of early detection.

Looking ahead, there are still several challenges (and opportunities) for making such techniques into a real clinical application. First, we may need a more pertinent taxonomy of specific emotions that signal concealed depression. 
Second, large-scale and correctly labeled FME databases (especially spontaneous ones) are in great need to build effective AI models, but are unfortunately not available due to privacy concerns. We are therefore in hope that the introduction of our new approach could contribute to solving this problem due to its privacy-preserving and self-diagnosis-enabling properties, which would allow the public to contribute data samples via a crowdsourcing manner. Third, innovation in more effective feature learning techniques and machine learning models is also a promising research area toward improving the performance of depression diagnosis.

\bibliographystyle{IEEEtran}
\bibliography{mer}

\begin{thebibliography}{10}
\providecommand{\url}[1]{#1}
\csname url@samestyle\endcsname
\providecommand{\newblock}{\relax}
\providecommand{\bibinfo}[2]{#2}
\providecommand{\BIBentrySTDinterwordspacing}{\spaceskip=0pt\relax}
\providecommand{\BIBentryALTinterwordstretchfactor}{4}
\providecommand{\BIBentryALTinterwordspacing}{\spaceskip=\fontdimen2\font plus
\BIBentryALTinterwordstretchfactor\fontdimen3\font minus
  \fontdimen4\font\relax}
\providecommand{\BIBforeignlanguage}[2]{{%
\expandafter\ifx\csname l@#1\endcsname\relax
\typeout{** WARNING: IEEEtran.bst: No hyphenation pattern has been}%
\typeout{** loaded for the language `#1'. Using the pattern for}%
\typeout{** the default language instead.}%
\else
\language=\csname l@#1\endcsname
\fi
#2}}
\providecommand{\BIBdecl}{\relax}
\BIBdecl

\bibitem{who}
``Depression,'' https://www.who.int/health-topics/depression.

\bibitem{nimh}
``Major depression,''
  https://www.nimh.nih.gov/health/statistics/major-depression.

\bibitem{afma}
Y.-H. Oh, J.~See, A.~C. Le~Ngo, R.~C. Phan, and V.~M. Baskaran, ``A survey of
  automatic facial micro-expression analysis: Databases, methods, and
  challenges,'' \emph{Frontiers in Psychology}, vol.~9, 2018.

\bibitem{ufe}
P.~Ekman, ``Universal and cultural differences in facial expression of
  emotion,'' in \emph{J. Cole (Ed.) Nebraska Symposium on Motivation, 1971},
  vol.~19.\hskip 1em plus 0.5em minus 0.4em\relax Lincoln University of
  Nebraska Press, 1972.

\bibitem{facs}
P.~Ekman and W.~V. Friesen, ``Facial action coding system,'' \emph{PsycTESTS
  Dataset}, 1978.

\bibitem{casme2}
W.-J. Yan, X.~Li, S.-J. Wang, G.~Zhao, Y.-J. Liu, Y.-H. Chen, and X.~Fu,
  ``Casme ii: An improved spontaneous micro-expression database and the
  baseline evaluation,'' \emph{PLoS ONE}, vol.~9, no.~1, 2014.

\bibitem{fee}
P.~Ekman, ``Facial expression and emotion,'' \emph{American Psychologist},
  vol.~48, p. 384–392, 1993.

\bibitem{joshi2022materials}
M.~L. Joshi and N.~Kanoongo, ``Depression detection using emotional artificial
  intelligence and machine learning: A closer review,'' \emph{Materials Today:
  Proceedings}, vol.~58, p. 217–226, 2022.

\bibitem{verma2022deep}
M.~Verma, S.~K. Vipparthi, and G.~Singh, ``Deep insights of learning based
  micro expression recognition: A perspective on promises, challenges and
  research needs,'' 2022.

\bibitem{li2022ieeetransactions}
Y.~Li, J.~Wei, Y.~Liu, J.~Kauttonen, and G.~Zhao, ``Deep learning for
  micro-expression recognition: A survey,'' \emph{IEEE Transactions on
  Affective Computing}, vol.~13, no.~4, pp. 2028--2046, 2022.

\bibitem{lee2022diagnostics}
Y.-S. Lee and W.-H. Park, ``Diagnosis of depressive disorder model on facial
  expression based on fast r-cnn,'' \emph{Diagnostics}, vol.~12, no.~2, p. 317,
  2022.

\bibitem{Abroms1981}
G.~M. Abroms, ``The diagnosis and treatment of hidden depression,''
  \emph{Psychiatric Quarterly}, no.~4, p. 235–241, 1981.

\bibitem{Surguladze2010}
S.~A. Surguladze, W.~El-Hage, T.~Dalgleish, J.~Radua, B.~Gohier, and M.~L.
  Phillips, ``Depression is associated with increased sensitivity to signals of
  disgust: A functional magnetic resonance imaging study,'' \emph{Journal of
  Psychiatric Research}, vol.~44, no.~14, p. 894–902, 2010.

\bibitem{lbp}
T.~Ojala, M.~Pietikäinen, and D.~Harwood, ``A comparative study of texture
  measures with classification based on featured distributions,'' \emph{Pattern
  Recognition}, vol.~29, no.~1, p. 51–59, 1996.

\bibitem{lbptop}
G.~Zhao and M.~Pietikainen, ``Dynamic texture recognition using local binary
  patterns with an application to facial expressions,'' \emph{IEEE Transactions
  on Pattern Analysis and Machine Intelligence}, vol.~29, no.~6, p. 915–928,
  2007.

\end{thebibliography}

\end{document}